\documentclass[10pt,twocolumn,letterpaper]{article}

\usepackage{iccv}
\usepackage{times}
\usepackage{epsfig}
\usepackage{graphicx}
\usepackage{amsmath}
\usepackage{amssymb}
\DeclareMathOperator*{\argmax}{argmax}
\usepackage{booktabs}
\usepackage{algorithm}
\usepackage{algorithmic}
\usepackage{alltt}
\usepackage{color}
\usepackage[dvipsnames]{xcolor,colortbl}
\usepackage{svg}
\usepackage{booktabs}
\usepackage{multicol,multirow}
\usepackage{makecell}

\usepackage[pagebackref=true,breaklinks=true,letterpaper=true,colorlinks,bookmarks=false]{hyperref}

\iccvfinalcopy 

\def\iccvPaperID{1213} 

\ificcvfinal\fi

\begin{document}

\title{Cumulative Spatial Knowledge Distillation for Vision Transformers}


\author{Borui Zhao$^{1}$ \quad Renjie Song$^{1}$ \quad  Jiajun Liang$^{1}$ \\
\vspace{-10pt}\\
$^{1}$MEGVII Technology\\
{\tt\small zhaoborui.gm@gmail.com, song.renjie@foxmail.com, liangjiajun@megvii.com} 
}

\maketitle
\ificcvfinal\thispagestyle{empty}\fi

\begin{abstract}
Distilling knowledge from convolutional neural networks~(CNNs) is \textbf{a double-edged sword} for vision transformers~(ViTs). It boosts the performance since the image-friendly local-inductive bias of CNN helps ViT learn faster and better, but leading to two problems:
(1)~Network designs of CNN and ViT are completely different, which leads to different semantic levels of intermediate features, making spatial-wise knowledge transfer methods~(\eg, feature mimicking) inefficient. (2)~Distilling knowledge from CNN limits the network convergence in the later training period since ViT's capability of integrating global information is suppressed by CNN's local-inductive-bias supervision.

To this end, we present Cumulative Spatial Knowledge Distillation~(CSKD). CSKD distills spatial-wise knowledge to all patch tokens of ViT from the corresponding spatial responses of CNN, without introducing intermediate features.
Furthermore, CSKD exploits a Cumulative Knowledge Fusion~(CKF) module, which introduces the global response of CNN and increasingly emphasizes its importance during the training. Applying CKF leverages CNN's local inductive bias in the early training period and gives full play to ViT's global capability in the later one. Extensive experiments and analysis on ImageNet-1k and downstream datasets demonstrate the superiority of our CSKD. Code will be publicly available.
\end{abstract}

\section{Introduction}
\label{sec:intro}

\begin{figure*}[th]
\centering
\includegraphics[width=0.98\textwidth]{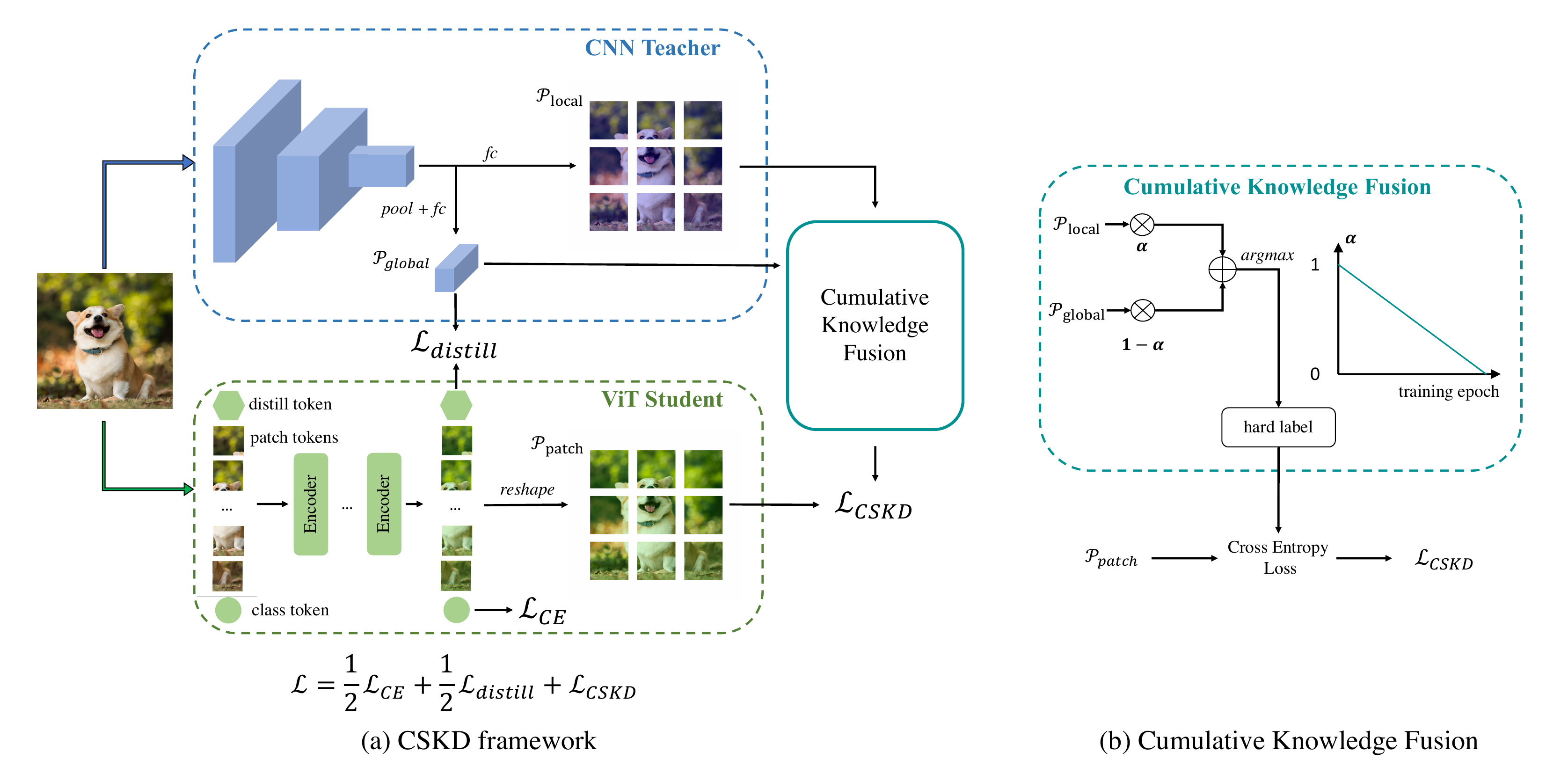}
\caption{Illustration of our Cumulative Spatial Knowledge Distillation~(CSKD).}
\label{fig:fig1}
\end{figure*}

Recently, Vision Transformers~(ViT) attract lots of attention in the computer vision community. Due to the powerful network capacity brought by the self-attention module, ViT achieves great performance on many vision tasks~(\eg, image recognition\cite{vit,liu2021swin} and object detection\cite{li2022exploring,carion2020end}). Knowledge distillation~(KD) is a widely-used technique, which improves the performance of a lightweight model~(the student) by transferring knowledge from a heavy one~(the teacher). DeiT\cite{deit} first applies this technique to ViT, which proposes to distill knowledge from the convolutional neural networks~(CNNs) for better training efficiency. The network design of CNN is image-friendly since the convolution operation is translation invariant, \ie, CNN has a local inductive bias which benefits the vision tasks. Thus, applying DeiT could help ViT converge faster and better for vision tasks. However, distilling knowledge from CNN is a double-edged sword. It introduces two problems, which are explained as follows.

Firstly, the network designs of CNN and ViT are different in three aspects: (1)~The ways to extend the receptive field, (2)~the ways to stack blocks, and (3)~the types of normalization layers. It makes intermediate features hard to align in terms of semantics. Thus, feature-based distillation methods could be inefficient due to the misalignment. The inefficiency is worthy of attention because the features can provide spatial-wise supervision with abundant knowledge. 

Secondly, CNN’s local inductive bias could hinder the full power of long-range dependencies and positional encoding of ViTs at convergence, although it helps ViT converge faster in the earlier stage~(which is also observed in \cite{chen2022dearkd}).
This phenomenon can be explained as follows: ViT's network capability is stronger than CNN's due to its powerful self-attention module. It could predict some hard training samples correctly that CNN can not handle. However, distilling knowledge from CNN's predictions provides negative supervision, leading ViT to predict wrong labels. With the progress of training, this problem is going to get worse.

Previous works make efforts to solve the problems above. (1) To alleviate the first one, DearKD\cite{chen2022dearkd} presents the MHCA module which modifies ViT's network architecture to align the intermediate features better. Despite the improved performance, extra module design is needed and ViT's capability of modeling the global relation is weakened.
(2) As for the second one, DearKD~\cite{chen2022dearkd} presents a two-stage learning strategy, and the distillation loss is disabled in the second stage to bypass the problem.
In CviT\cite{ren2022co}, multiple networks with various inductive biases are used as teachers to provide more abundant knowledge, so that ViT's capability could be less suppressed. Both works boost the distillation performance. However, the former degrades into a baseline training scheme in the later period because it doesn't apply knowledge distillation, and the latter makes the training process more computationally intensive.

In this paper, we present Cumulative Spatial Knowledge Distillation~(CSKD) to alleviate the aforementioned problems in a simple but effective way. (1) CSKD generates spatial classification predictions from CNN's last features and uses them as the supervision signals of the corresponding patch tokens of ViT~(as shown in Figure~\ref{fig:fig1} (a)). In this way, spatial-wise knowledge is transferred without introducing intermediate features. Thus, the first problem can be alleviated successfully. (2) CSKD exploits a Cumulative Knowledge Fusion~(CKF) module which fuses the local and the global responses and increasingly emphasizes the importance of the global ones with the progress of training.~(as shown in Figure~\ref{fig:fig1} (b)). In this way, CKF leverages the local inductive bias for fast convergence in the early period and gives full play to ViT's network capability in the later one. Experimental results on ImageNet-1k and downstream datasets demonstrate the effectiveness of our CSKD. We further provide visualizations to validate that CSKD leverages ViT's network capability better.

Our contributions are summarized as follows.
\begin{itemize}
    \item {Two obstacles to distilling knowledge from CNN to ViT are discussed: (1) the spatial-wise knowledge transfer is inefficient and (2) the network convergence is limited in the later training period.}
    \item {We present Cumulative Spatial Knowledge Distillation~(CSKD) to alleviate the two problems in a simple yet effective way. CSKD transfers spatial-wise knowledge without introducing intermediate features and increasingly emphasizes the importance of global responses for better supervision.}
    \item {Extensive experiments validate the effectiveness of our CSKD. We further provide analysis and visualizations, demonstrating that CSKD gives full play to ViT's powerful network capability.}
\end{itemize}

\section{Related Work}
\subsection{ConvNets and Vision Transformers}
Convolutional neural networks~(CNN) have been the mainstream on computer vision tasks since \cite{lenet} firstly presents the convolution operation. Convolution is translation invariant~(\ie, with an image-friendly local inductive bias) since the convolution filters are replicated across the entire visual field and share the same parameters. Many works further investigate the network design of CNN. AlexNet\cite{alexnet} firstly propose a deep neural network that performs well on the large-scale ImageNet-1k dataset. GoogLeNet\cite{googlenet} and ResNet\cite{resnet} further explore deeper and more effective architectures. Meanwhile, MobileNet\cite{mobilenet,mobilenetv2} and ShuffleNet\cite{shufflenet,shufflenetv2} series design lightweight networks to efficiently deploy CNNs on mobile devices.

Transformers are proposed by \cite{vaswani2017attention} for the machine translation task. The powerful self-attention module helps the network to model long-distance relations, making transformers the state of the art in many natural language processing~(NLP) tasks.
Vision transformers~(ViTs) are firstly proposed by \cite{vit}, which proves that the powerful global self-attention module can also boost performance for vision tasks. Recently, methods introducing more image-friendly inductive bias are proposed for better performance. Swin Transformer\cite{liu2021swin} presents a shifted windowing scheme to limit self-attention computation to local windows. PVT\cite{wang2021pyramid} employs feature pyramid to reduce the computations and benefits dense-prediction vision tasks~(\eg, object detection). T2T-ViT\cite{yuan2021tokens} recursively aggregates neighboring tokens of an overlapping sliding window, improving the training efficiency. The above works demonstrate that image-friendly design is still important despite ViT's powerful network capacity.
\subsection{Knowledge Distillation}
\textbf{CNN}. Knowledge distillation~(KD) is proven efficient for transferring knowledge from heavy models~(teacher) to lightweight ones~(student), achieving great success on many vision tasks. KD is firstly proposed by \cite{kd}, which introduces the teacher's predictions as ``soft" labels and uses them to supervise the student. Besides such logit-based methods\cite{kd,takd,eskd,zhao2022decoupled}, feature-mimicking methods are proposed to leverage the spatial-wise knowledge, since the intermediate features have spatial dimensions. FitNets\cite{fitnets} firstly distill hints from intermediate features. Many works \cite{at,ofd,rkd,crd} further delve into distilling feature-based knowledge from different perspectives. There are also many works\cite{detmimick1,detmimick2,liu2019structured,zhang2019fast} applying KD to more vision tasks(\eg, detection and segmentation).

\textbf{ViT}. To benefit the ViT community with KD, DeiT\cite{deit} proposes to distill knowledge from CNNs to vision transformers, using CNN's local inductive bias for better training efficiency. After that, DearKD\cite{chen2022dearkd} proposes a two-stage learning framework, distilling knowledge from CNN's intermediate features only at the early stage. DearKD learns stronger global representations yet needs extra designs for feature alignment and training schemes. CivT\cite{ren2022co} focuses on distilling knowledge from more models with different inductive biases. It introduces both convolution and involution\cite{li2021involution} networks as teachers, achieving better performance. Previous works prove the potential of distilling knowledge from different-inductive-bias networks, which motivates us to explore further.

\section{Distilling Knowledge from CNN is a Double-edge Sword}

\subsection{Recap of Vision Transformer Distillation}
\paragraph{Vision Transformers}
Inspired by NLP successes, vision transformers~(ViTs) are firstly proposed by \cite{vit}. ViT takes an image as several ``$16\times 16 words$" and utilizes transformers for the image recognition task. Firstly, an input image~(\eg, with $224 \times 224$ resolution and $K$ channels) is split into $N$ discrete patches with patch size $P$. Each patch is a vector with $KP^2$ channels. Thus we have $N=224^2/P^2$. Then, every patch is mapped to $D$ dimensions as patch embeddings~(with position embeddings to retain positional information). An extra learnable embedding named ``class token" is introduced for the final classification prediction. ViT feeds these tokens to stacked transformer encoders to model global relations. Every transformer is composed of multi-head self-attention modules~(MHSA), MLP blocks, and layer normalization~(LN) blocks. At last, ViT takes the output class token from the last encoder as the features for classification and feeds them to a fully-connected layer for the final predictions.

\paragraph{DeiT}
To further improve the training efficiency, DeiT first proposes to train ViT via transferring knowledge from CNN. It utilizes CNN's image-friendly inductive bias for better supervision. Concretely, DeiT introduces a new ``distillation token" to ViT. The distillation token takes the output logits of the CNN teacher network as the supervision signal. The corresponding loss function $\mathcal{L}_{\text{distill}}$ could be either KL-Divergence~(like the traditional KD\cite{kd}) or cross-entropy loss~(the target label is the hard decision of the teacher). The training loss of DeiT composes of the default cross-entropy loss supervised by the target label~(denoted as $\mathcal{L}_{\text{CE}}$) and the aforementioned distillation loss, Eqn.~(\ref{eq1}).
\begin{equation}
    \mathcal{L} = \frac{1}{2}\mathcal{L}_{\text{CE}} + \frac{1}{2} \mathcal{L}_{\text{distill}}
    \label{eq1}
\end{equation}
As for the inference phase, DeiT takes the average of class token output and distillation token output as the final output. DeiT~\cite{deit} reveals two phenomena. (1) CNNs are better teachers than ViTs because the inductive bias of CNNs provides more image-friendly guidance. (2) Hard distillation is better than soft distillation~\footnote{Thus, we simply use hard distillation by default in the rest of this paper.}.

\subsection{A Double-edged Sword}

DeiT achieves much better training efficiency by distilling knowledge from CNN. However, distilling knowledge from a teacher with such a different inductive bias is a \textit{double-edged sword}.
On the bright side, CNN's local inductive bias helps ViT converge faster and better. On the dark side, the difference in network architectures and network capabilities limits the distillation potential.

\paragraph{Different network architectures.} CNN's network design is completely different from ViT's. The main differences are as follows: (1) The ways to \textit{extend the receptive field} are different between CNN and ViT. CNN mainly extends its receptive field by downsampling the features step by step~(\eg, the feature map of the first stage is $16\times16$ and the second is $8\times8$ for spatial dimensions). Stacking spatial-wise convolution~(\eg, $3\times 3$) also provides assistance. ViT extends the network receptive field by introducing self-attention. Self-attention makes ViT have a global receptive field from the first stage. (2) The ways to \textit{stack blocks} are also different. All blocks share the same architectures in ViT, yet CNN's stage designs are different from each other~(\eg, the channel dimensions and the number of bottleneck modules are different among different stages in a ResNet50~\cite{resnet} network). (3) The \textit{types of normalization layers} are different between CNN and ViT. CNN always uses batch normalization~\cite{ioffe2015batch} and group normalization~\cite{wu2018group} while ViT utilizes layer normalization~\cite{ba2016layer}.

All the differences above make the intermediate features of CNNs and ViTs have different semantic levels. It makes the intermediate features hard to align. Thus, feature-based distillation methods~(\ie, feature-mimicking) could be inefficient due to the misalignment. The inefficiency should be attached much importance to because the features can provide spatial-wise supervision with abundant knowledge. To transfer spatial-wise knowledge, DearKD\cite{chen2022dearkd} presents the MHCA module cast in the ViT network architecture, achieving better performance than DeiT. However, it modifies the network structure and limits the capability of global relation modeling.

\paragraph{Different network capability.}
As is well-known, the excellent performance of ViT is owing to the global receptive field provided by the self-attention module. Relatively speaking, the network capability of CNN is not \textit{that powerful} since there is no module having the global-relation modeling ability in CNN's architecture.
However, stronger teachers are favored in the knowledge distillation community since a better-performance network could transfer more valuable knowledge~\footnote{There are also some works\cite{tfkd} investigating how to distill knowledge from less-powerful networks. Yet distilling knowledge from better ones is still the mainstream.}.
Distilling knowledge from CNN to ViT \textit{violates the traditional practice} since the CNN teacher is less powerful. Thus, when the student network is converged, its performance could surpass the teacher's. At this point, the supervision signal of the teacher could be harmful. Previous works attempt to alleviate this problem from different perspectives. DearKD\cite{chen2022dearkd} presents a two-stage learning strategy, deprecating the distillation term in the second stage. It achieves better performance yet wastes the training time in the second stage~(\ie, without applying knowledge distillation). CviT\cite{ren2022co} uses multiple networks as teachers whose network designs are different ~(\eg, CNN and INN~\cite{li2021involution}) to provide knowledge of various inductive biases. It enriches the knowledge yet makes the training more computationally intensive because of the introduction of multiple teachers.

In summary, distilling knowledge from CNN to ViT is a double-edged sword. The local inductive bias of CNN helps ViT learn image-friendly knowledge. However, the difference in architectural design makes the intermediate features hard to align, limiting the efficiency of spatial-wise knowledge transfer. Meanwhile, the gap in network capability makes CNN's supervision harmful to some degree in the later training period.

\section{Cumulative Spatial Knowledge Distillation}

Two problems of distilling knowledge from CNN to ViT are revealed in the previous section. Driven by the analysis, we present a simple but effective method named Cumulative Spatial Knowledge Distillation~(CSKD). CSKD transfers spatial-wise knowledge without introducing intermediate features and alleviates the capability mismatch issue by cumulative learning global-wise knowledge.

\paragraph{Notations}
We denote the input image as $\mathcal{X}$ and the target label as $\mathcal{Y}$. The CNN teacher and the ViT student are denoted as $\mathcal{T}$ and $\mathcal{S}$, respectively. C denotes the number of classes for the image recognition task. For simplification, we omit the batch size of training samples.

\subsection{Spatial-wise Knowledge Transfer}

Firstly, we generate dense classification predictions to provide spatial-wise knowledge. Dense predictions could transfer spatial-wise information and avoid complex feature aligners like \cite{chen2022dearkd}. Ways to generate dense predictions are different between CNN and ViT. As for CNN, the last feature map~(with $H \times W$ resolution) is fed to a global average pooling layer and a final fully-connected layer~(classifier) to generate the global output logits~(denoted as $\mathcal{P}_{global}^{\mathcal{T}} \in \mathbb{R}^{1\times C}$). We remove the pooling operation and take the features of each position as an \textit{independent sample}. Then, we feed them to the classifier, generating dense predictions of every corresponding position. We denote the dense predictions from CNNs as $\mathcal{P}^{\mathcal{T}} \in \mathbb{R}^{1\times H \times W \times C}$. As for ViTs, we take every patch token from the last transformer encoder as the final feature of the corresponding position. We further feed the final features to the fully-connected classifier to generate dense predictions. Dense predictions~\footnote{The shape of ViT's dense predictions is (1, N, C) by default. Details about aligning them to the same shape of CNN's predictions are attached to the supplement.} from ViTs are denoted as $\mathcal{P}^{\mathcal{S}} \in \mathbb{R}^{1\times H \times W \times C}$~\footnote{In Figure~\ref{fig:fig1}, $\mathcal{P}^{\mathcal{T}}$ and $\mathcal{P}^{\mathcal{S}}$ are rewritten as $\mathcal{P}_{\text{local}}$ and $\mathcal{P}_{\text{patch}}$ respectively for easy understanding.}. Since every patch token is embedded with the corresponding sub-image and position information, regarding them as spatial dense predictions is reasonable. At last, we take the hard decisions~(\ie, \textit{argmax}) of dense predictions from CNN as target labels~(denoted as $\mathcal{Y}^{\mathcal{T}}$) and use cross-entropy~(CE) as the loss function for spatial knowledge transfer.

\subsection{Cumulative Knowledge Fusion}

To eliminate the effect of capacity mismatch, we introduce a Cumulative Knowledge Fusion~(CKF) module. According to \cite{chen2022dearkd}, CNN helps ViT converge better in the early stage of training due to the different inductive biases, but limits the network convergence in the later period. As aforementioned, DearKD presents a two-stage distillation framework and disables the knowledge distillation to mitigate this problem. In contrast, we believe that CNN could also help ViT learn better representations in the later period. The key factor to achieve this goal is \textit{using better supervision signals}. In the later training period, each patch token of ViT has already modeled global relations among other tokens. Thus, supervising the tokens with global information instead of local information could help them leverage their capability and learn better representations. Thus, we present the Cumulative Knowledge Fusion module~(CKF). As illustrated in Figure~\ref{fig:fig1}~(b), CKF fuses CNN's global~($\mathcal{P}_{global}^{\mathcal{T}}$) and local~($\mathcal{P}^{\mathcal{T}}$) responses during the entire training process. It prefers local ones in the early stage and global ones in the later stage. CKF generates the dense target labels according to :
\begin{equation}
    \mathcal{Y}^\mathcal{T} = \mathop{\argmax}_{c}{\Big( \alpha \mathcal{P}^{\mathcal{T}} + (1 - \alpha) \mathcal{P}_{global}^{\mathcal{T}}\Big)},
\label{eq:yt}
\end{equation}
where $\alpha$ is a coefficient that  negatively correlates with the training duration(\eg, $\alpha = 1 - t/t_{max}$, where $t$ denotes the current epoch and $t_{max}$ denotes the number of total epochs). Thus, the loss function of our CSKD can be written as :

\begin{equation}
\begin{split}
    \mathcal{L}_{\text{CSKD}} &= \text{CE}(\mathcal{P}^{\mathcal{S}}, \mathcal{Y}^{\mathcal{T}})
\end{split}
\label{eq:cskd}
\end{equation}
As Eqn.~(\ref{eq:yt}) suggests, the more training time increases the more global knowledge is transferred. Even if there is a capacity mismatch between CNNs and ViTs, supervising local patches with global knowledge could alleviate the problem and further improve the distillation performance. We explore visualizations in Section~\ref{subsec:vis} to verify it.

In summary, CSKD transfers spatial-wise knowledge through the patch tokens of ViT without introducing feature mimicking. It also benefits ViT in the later training period by increasingly emphasizing the importance of global responses. We keep the cross-entropy loss and the distillation loss in DeiT~\cite{deit}. The total loss can be written as:~\footnote{There is a default hyperparameter to weight $\mathcal{L}_{\text{CSKD}}$. Since we conduct ablation studies on this hyperparameter and validate the robustness among different values in Table~\ref{tab:abl4}, it is omitted for simplification.}
\begin{equation}
    \mathcal{L} = \frac{1}{2} \mathcal{L}_{\text{CE}} + \frac{1}{2} \mathcal{L}_{\text{distill}} + \mathcal{L}_{\text{CSKD}}
\end{equation}
Algorithm~\ref{algo} provides CSKD's PyTorch-style pseudocode.

\begin{algorithm}[t]
    \caption{Pseudocode of CSKD in PyTorch.}
    \label{algo}
    \footnotesize
    \begin{alltt}
    \color{BlueGreen}
# f_cnn: CNN features before global pooling
# f_vit: ViT features of all patch tokens
# cls_cnn: classifier of the teacher
# cls_vit: classifier of the student
# T: the temperature for KD
\end{alltt}
\begin{alltt}
\color{BlueGreen}
# get CNN's global & dense predictions \color{Black}
b, c1, h, w = f_cnn.shape
p_global = cls_cnn(f_cnn.mean(dim=[2,3]))
f_cnn = f_cnn.permute(0, 2, 3, 1)
f_cnn = f_cnn.reshape(b, -1) #(b*h*w, c)
p_cnn = cls_cnn(f_cnn)
\color{BlueGreen}
# get ViT's dense predictions \color{Black}
b, c2, hw = f_vit.shape
f_vit = f_vit.permute(0, 2, 1)
f_vit = f_vit.reshape(b, -1)
p_vit = cls_vit(f_vit)
\color{BlueGreen}
# calculate loss_cskd \color{Black}
alpha = 1 - epoch / max_epoch
p_fusion = alpha*p_cnn + (1-alpha)*p_global
target = p_fusion.argmax(dim=1)

loss_cskd = CrossEntropyLoss(p_vit, target)
\end{alltt}
\end{algorithm}

\section{Experiments}
\label{sec:exp}
In this section, we mainly conduct experiments on ImageNet-1k\cite{imagenet} to evaluate CSKD. Besides, experiments on transfer learning to downstream tasks validate the generalization ability of CSKD. We also provide ablation studies and visualizations for better understanding.
\subsection{Implementation Details}
\label{subsec:impl}
Our implementation \textit{exactly} follows the practice in DeiT\cite{deit}. Concretely, we use a RegNetY-16GF CNN model as the teacher, which achieves 82.9\% top-1 accuracy on the ImageNet-1k validation set. All models are trained for 300 epochs with 1024 batch size. We use AdamW as the optimizer with a 0.001 initial learning rate, and cosine decay as the learning scheduler~(with a warmup for the first 5 epochs). The weight decay is set as 0.05. Input images are at resolution $224\times224$. Mixup\cite{zhang2017mixup}, Cutmix\cite{yun2019cutmix}, and Auto-Augment\cite{autoaug} are utilized for data augmentation, as the same as DeiT\cite{deit}. Stochastic depth\cite{huang2016deep} is not employed since we already apply knowledge distillation which is stronger regularization.

As the same as DeiT\cite{deit}~\footnote{Since we only consider the  knowledge distillation setting, all ``DeiT" represent the distilled ones for simplification in the rest of the paper.}, we also train tiny, small, and base models named CSKD-Ti, CSKD-S, and CSKD-B, respectively. Table~\ref{tab:model_info} summarizes the model information.
\begin{table}[ht]
    \centering
    \begin{small}
    \begin{tabular}{c|cccc}
        model & \#heads & \#layers & \#params & throughput \\
        \Xhline{3\arrayrulewidth}
        CSKD-Ti & 3 & 12 & 5M & 2536 \\
        CSKD-S & 6 & 12 & 22M & 940 \\
        CSKD-B & 12 & 12 & 86M & 292 \\
        
    \end{tabular}
    \caption{Variants of the model architecture. The larger model DeiT-B and CSKD-B have the same architecture as the ViT-B. Embedding dimensions and the number of heads are different among different variants. Dimension per head is always 64. The throughput~(img/sec) is measured at resolution $224\times224$.}
    \label{tab:model_info}
    \end{small}
\end{table}

\begin{table}[th]
    \centering
    \begin{small}
    \begin{tabular}{ccc|cc}
\multicolumn{1}{c|}{model}    & \#params & \begin{tabular}[c]{@{}c@{}}image\\ size\end{tabular} & \begin{tabular}[c]{@{}c@{}}val\\ top-1\end{tabular} & \begin{tabular}[c]{@{}c@{}}val-real\\ top-1\end{tabular} \\ \Xhline{3\arrayrulewidth}
\multicolumn{5}{c}{\textit{CNNs}} \\ 
\hline
\multicolumn{1}{c|}{ResNet18\cite{resnet}} & 12M & $224^2$ & 69.8 & 77.3 \\
\multicolumn{1}{c|}{ResNet50\cite{resnet}} & 25M & $224^2$ & 76.2 & 82.5 \\  
\multicolumn{1}{c|}{ResNet101\cite{resnet}} & 45M & $224^2$ & 77.4 & 83.7 \\  
\multicolumn{1}{c|}{ResNet152\cite{resnet}} & 60M & $224^2$ & 78.3 & 84.1 \\
\hline
\multicolumn{1}{c|}{RegNetY-4GF*\cite{radosavovic2020designing}} & 21M & $224^2$ & 80.0 & 86.4 \\  
\multicolumn{1}{c|}{RegNetY-8GF*\cite{radosavovic2020designing}} & 39M & $224^2$ & 81.7 & 87.4 \\ 
\multicolumn{1}{c|}{RegNetY-16GF*\cite{radosavovic2020designing}} & 84M & $224^2$ & 82.9 & 88.1 \\ 
\hline
\multicolumn{1}{c|}{EffiNet-B0\cite{tan2019efficientnet}} & 5M & $224^2$ & 77.1 & 84.1 \\
\multicolumn{1}{c|}{EffiNet-B3\cite{tan2019efficientnet}} & 12M & $224^2$ & 81.6 & 86.8 \\
\multicolumn{1}{c|}{EffiNet-B6\cite{tan2019efficientnet}} & 43M & $224^2$ & 84.0 & 88.8 \\
\hline
\multicolumn{5}{c}{\textit{ViTs}} \\ 
\hline
\multicolumn{1}{c|}{ViT-B/16\cite{vit}} & 86M & $384^2$ & 77.9 & 83.6 \\
\multicolumn{1}{c|}{ViT-L/16\cite{vit}} & 307M & $384^2$ & 76.5 & 82.2 \\
\hline
\multicolumn{1}{c|}{DeiT-Ti\cite{deit}} & 6M & $224^2$ & 74.5 & 82.1 \\
\multicolumn{1}{c|}{DeiT-S\cite{deit}} & 22M & $224^2$ & 81.2 & 86.8 \\
\multicolumn{1}{c|}{DeiT-B\cite{deit}} & 87M & $224^2$ & 83.4 & 88.3 \\
\hline
\multicolumn{1}{c|}{DearKD-Ti\cite{chen2022dearkd}} & 5M & $224^2$ & 74.8 & - \\
\multicolumn{1}{c|}{DearKD-S\cite{chen2022dearkd}} & 22M & $224^2$ & 81.5 & - \\
\multicolumn{1}{c|}{DearKD-B\cite{chen2022dearkd}} & 86M & $224^2$ & 83.6 & - \\
\hline
\multicolumn{1}{c|}{\textbf{CSKD-Ti}} & 6M & $224^2$ & \textbf{76.3} & \textbf{83.6} \\
\multicolumn{1}{c|}{\textbf{CSKD-S}} & 22M & $224^2$ &   \textbf{82.3} & \textbf{87.9} \\
\multicolumn{1}{c|}{\textbf{CSKD-B}} & 87M & $224^2$ & \textbf{83.8} & \textbf{88.6} \\

    \end{tabular}
    \caption{\textbf{Results on ImageNet-1k}. * means that RegNets are optimized with similar optimization procedures as DeiT, serving as teachers for DeiT and our CSKD. ``val-real" represents the results of the ImageNet Real validation set.}
    \label{tab:imagenet}
    \end{small}
\end{table}

\subsection{Results on ImageNet-1k}
We mainly evaluate our CSKD on the ImageNet-1k\cite{imagenet} dataset.
ImageNet-1k is a well-known large-scale dataset that consists of 1000 classes. The training set contains 1.28 million images and the validation set contains 50k images. We also use ImageNet Real\cite{imagenetreal} for further evaluation, which corrects some wrong labels of the validation set.

Table~\ref{tab:imagenet} reports the comparison results on ImageNet-1k and demonstrates the effectiveness of CSKD. CSKD-Ti, CSKD-S, and CSKD-B improve the performance by $+1.8\%$, $+1.1\%$, and $+0.4\%$ respectively. And the results on ImageNet Real also verify the performance gain. 
Besides, we train CSKD-Ti for a longer time~(\ie, 1000 epochs). As shown in Table~\ref{tab:1k_epoch}, CSKD-Ti-1k achieves 78.1\% top-1 accuracy, surpassing the baseline by a large margin, further proving the effectiveness.

\begin{table}[ht]
    \centering
    \begin{small}
    \begin{tabular}{c|ccc}
        model & DeiT-Ti-1k & DearKD-Ti-1k & CSKD-Ti-1k \\
        \Xhline{3\arrayrulewidth}
        val top-1 & 76.6 & 77.0 & \textbf{78.1} 
    \end{tabular}
    \caption{We train our CSKD-Ti for \textit{1000 epochs}~(CSKD-Ti-1k). CSKD-Ti achieves 78.1\% top-1 accuracy, surpassing DeiT and DearKD by a large margin.}
    \label{tab:1k_epoch}
    \end{small}
\end{table}

\subsection{Transfer Learning to Downstream Datasets}
We conduct experiments on transfer learning to downstream datasets. CIFAR10\cite{cifar}, CIFAR100\cite{cifar}, Cars\cite{krause20133d} and iNat19 are considered. Table~\ref{tab:transfer_dataset} reports the dataset information. As shown in Table~\ref{tab:transfer}, our CSKD achieves significantly better transfer performances than DeiT and DearKD on all the downstream datasets. Especially for the challenging iNat19 fine-grained classification task, CSKD-Ti, CSKD-S, and CSKD-B surpass the corresponding baselines by $+1.1\%$, $+0.8\%$, and $+0.4\%$ respectively.

\begin{table}[ht]
    \centering
    \begin{small}
    \begin{tabular}{c|ccc}
        dataset & training size & val size & \#classes  \\
        \Xhline{3\arrayrulewidth}
        CIFAR10 & 50,000 & 10,000 & 10 \\
        CIFAR100 & 50,000 & 10,000 & 100 \\
        Stanford Cars & 8,144 & 8,041 & 196 \\
        iNaturalist 2019 & 265,240 & 3,030 & 1,010
    \end{tabular}
    \caption{Downstream dataset information for transfer learning.}
    \label{tab:transfer_dataset}
    \end{small}
\end{table}

\begin{table}[ht]
\centering
\begin{small}
\begin{tabular}{c|cccc}
  & CIFAR10 & CIFAR100 & Cars & iNat19   \\
 \Xhline{3\arrayrulewidth}
 DeiT-Ti* & 98.1 & 86.1 & 92.1 & 76.6 \\
 DearKD-Ti & 97.5 & 85.7 &  89.0 & - \\
 CSKD-Ti & \textbf{98.5} & \textbf{87.0} & \textbf{93.1} & \textbf{77.7} \\
 \hline
 DeiT-S* & 98.7 & 89.2 & 91.7 & 80.9 \\
 DearKD-S & 98.4 & 89.3 & 91.3 & - \\
 CSKD-S & \textbf{99.1} & \textbf{90.3} & \textbf{93.7} & \textbf{81.7} \\
 \hline
 DeiT-B* & 99.1 & 91.3 & 92.9 & 82.1 \\
 DearKD-B & 99.2 & 91.1 & 92.7 & - \\
 CSKD-B & \textbf{99.3} & \textbf{91.4} & \textbf{94.0} & \textbf{82.5} \\
\end{tabular}
\caption{\textbf{Transfer performance} in downstream tasks. * represents that the results are based on our implementation.}
\label{tab:transfer}
\end{small}
\end{table}

\subsection{Ablation Study}
We ablate key elements of our design in CSKD. All models are trained based on CSKD-B following the settings in Section~\ref{subsec:impl}. First, we investigate how distilling spatial-wise knowledge and CKF module work. Table~\ref{tab:abl1} reports the results of (1) the DeiT baseline and our CSKD that distills knowledge from (2) the local responses, (3) the global ones, (4) combining both by a simple summation, and (5) combining the both with CKF. The results indicate that: distilling spatial-wise knowledge can improve the performance~(81.2\% \vs 81.6\%). Moreover, introducing CKF significantly improves the performance from 81.5\%~(simply combining global and local responses) to 82.3\%.

\begin{table}[ht]
\centering
\begin{small}
\begin{tabular}{ccc|c}
local & global &  CKF & val top-1 \\
\Xhline{3\arrayrulewidth}
\space &\space & \space & 81.2 \\
\checkmark & \space & \space & 81.6\textcolor{PineGreen}{(+0.4)} \\
\space &\checkmark & \space & 81.4\textcolor{PineGreen}{(+0.2)} \\
\checkmark &\checkmark & \space & 81.5\textcolor{PineGreen}{(+0.3)} \\
\checkmark &\checkmark & \checkmark & 82.3\textcolor{PineGreen}{(+1.1)}
\end{tabular}
\caption{Ablation study on distilling knowledge from CNN's local and global responses. The great effect of our CKF is also verified~(81.5\% \vs 82.3).}
\label{tab:abl1}
\end{small}
\end{table}

We further explore different loss types. In traditional knowledge distillation methods, KL-Divergence is the commonly used loss function for logit-based distillation~(\ie, soft distillation) instead of the cross-entropy loss~(\ie, hard distillation). Results in Table~\ref{tab:abl2} demonstrate that hard distillation is more suitable for distilling knowledge from CNN to ViT, while the difference is marginal between them~(82.3\% \vs 82.1\%).

\begin{table}[h]
\centering
\begin{small}
\begin{tabular}{c|ccc}
distillation type & hard & soft & soft+hard \\
\Xhline{3\arrayrulewidth}
val top-1 & 82.3 & 82.1 & 82.2 \\
\end{tabular}
\caption{Ablations on different distillation loss types.}
\label{tab:abl2}
\end{small}
\end{table}

We study different strategies for decaying $\alpha$ in the CKF module. We use linear decay, cosine decay, and parabolic decay for comparison. Results in Table~\ref{tab:abl3} show that our CKF is insensitive to the decaying strategy since the results are pretty much the same~(82.3\% \vs 82.2\%).
\begin{table}[h]
\centering
\begin{small}
\begin{tabular}{c|c}
strategy & val top-1 \\
\Xhline{3\arrayrulewidth}
$1 - t/t_{max}$ & 82.3 \\
$\text{cos}(\pi/2 \cdot t/t_{max})$ & 82.2 \\
$(1 - t/t_{max})^2$ & 82.2
\end{tabular}
\caption{Different strategies for decaying $\alpha$.}
\label{tab:abl3}
\end{small}
\end{table}

\begin{table}[h]
\centering
\begin{small}
\begin{tabular}{c|ccc}
loss weight & 0.5 & 1.0 & 2.0 \\
\Xhline{3\arrayrulewidth}
val top-1 & 82.1 & 82.3 & 82.1 \\
\end{tabular}
\caption{Applying CSKD with different loss weight.}
\label{tab:abl4}
\end{small}
\end{table}

Lastly, we conduct experiments on tuning loss weights for CSKD. The results in Table~\ref{tab:abl4} indicate that weighing CSKD in a range of $\{0.5, 2.0\}$ achieves similar performance and 1.0 is the best one. In summary, the ablation studies above prove the effectiveness of CSKD and CKF, and also reveal that CSKD is hyperparameter insensitive.

\begin{figure}[ht]
    \centering
    \includegraphics[width=0.48\textwidth]{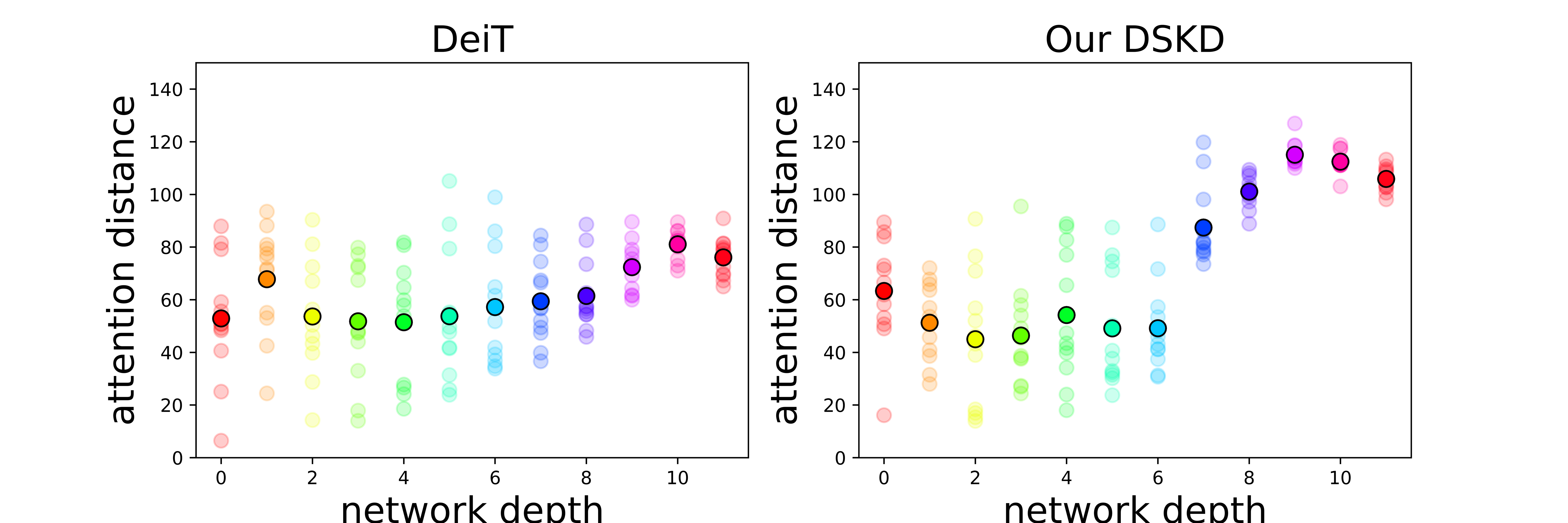}
    \caption{\textbf{Mean attention distance} of DeiT-B and CSKD-B. The final stages of CSKD have significantly larger attention distances than DeiT, which suggests that CSKD \textbf{leverages the global capacity of ViTs in a better way}.}
    \label{fig:attn_dist}
\end{figure}

\begin{figure}[ht]
    \centering
    \includegraphics[width=0.45\textwidth]{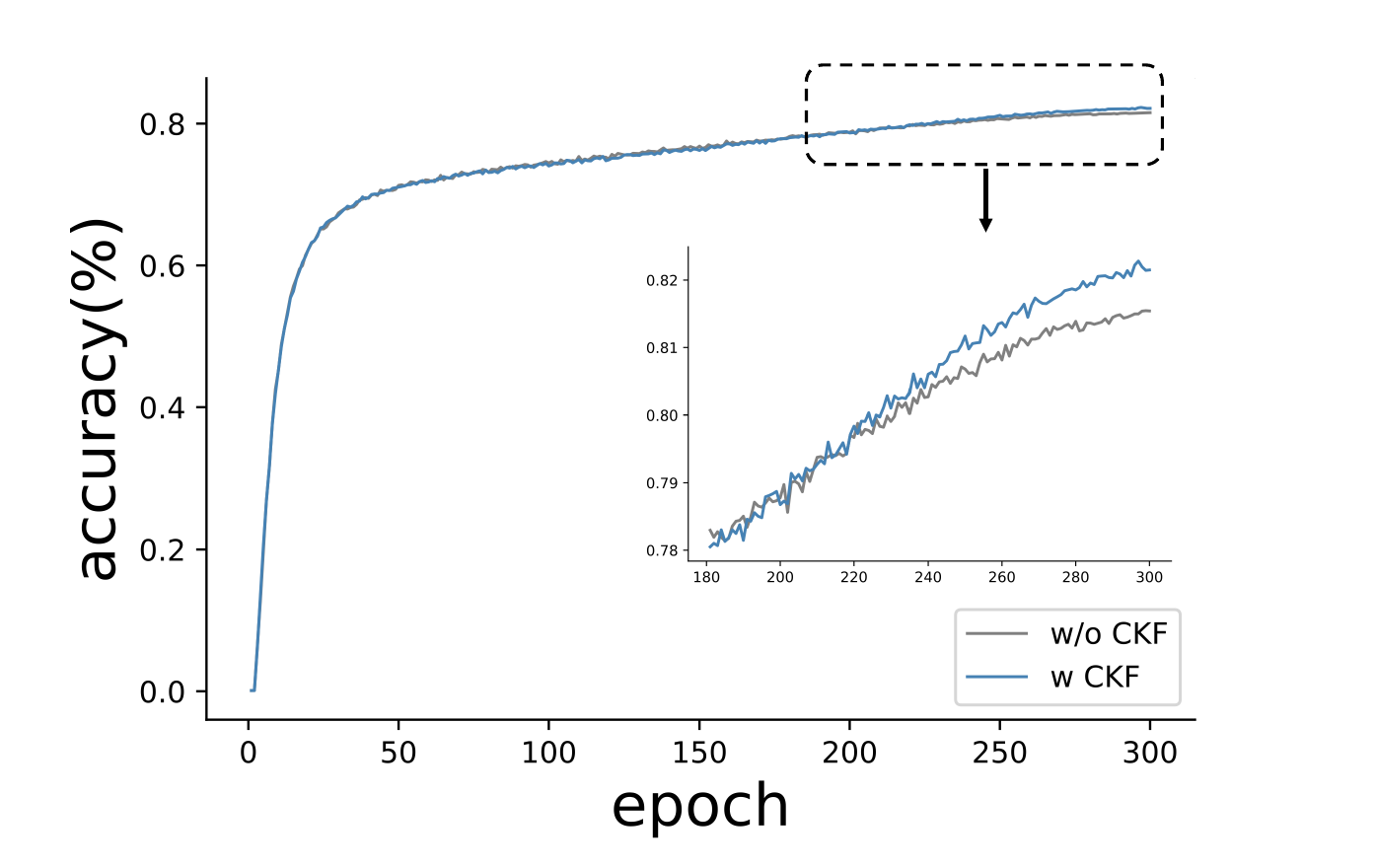}
    \caption{Validation accuracy dynamics. It indicates that CKF mainly \textbf{works in the later training period}.}
    \label{fig:curve}
\end{figure}

\begin{figure}[ht]
\centering	
\includegraphics[width=0.22\textwidth]{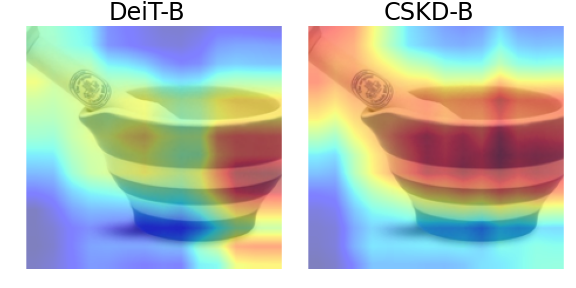}
\includegraphics[width=0.22\textwidth]{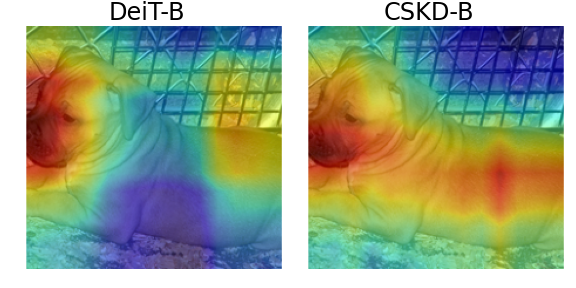}
\includegraphics[width=0.22\textwidth]{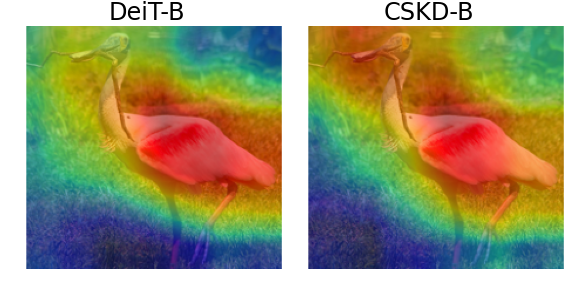}
\includegraphics[width=0.22\textwidth]{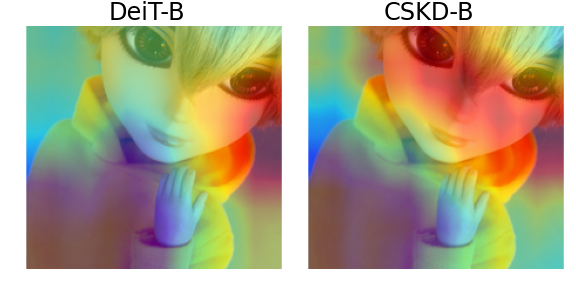}
\caption{Attention Heatmaps from DeiT-B and our CSKD-B. CSKD-B focuses more attention on the salient object.}
\label{fig:attnmap}
\end{figure}

\begin{figure*}[t!]
    \centering
        \includegraphics[width=0.43\textwidth]{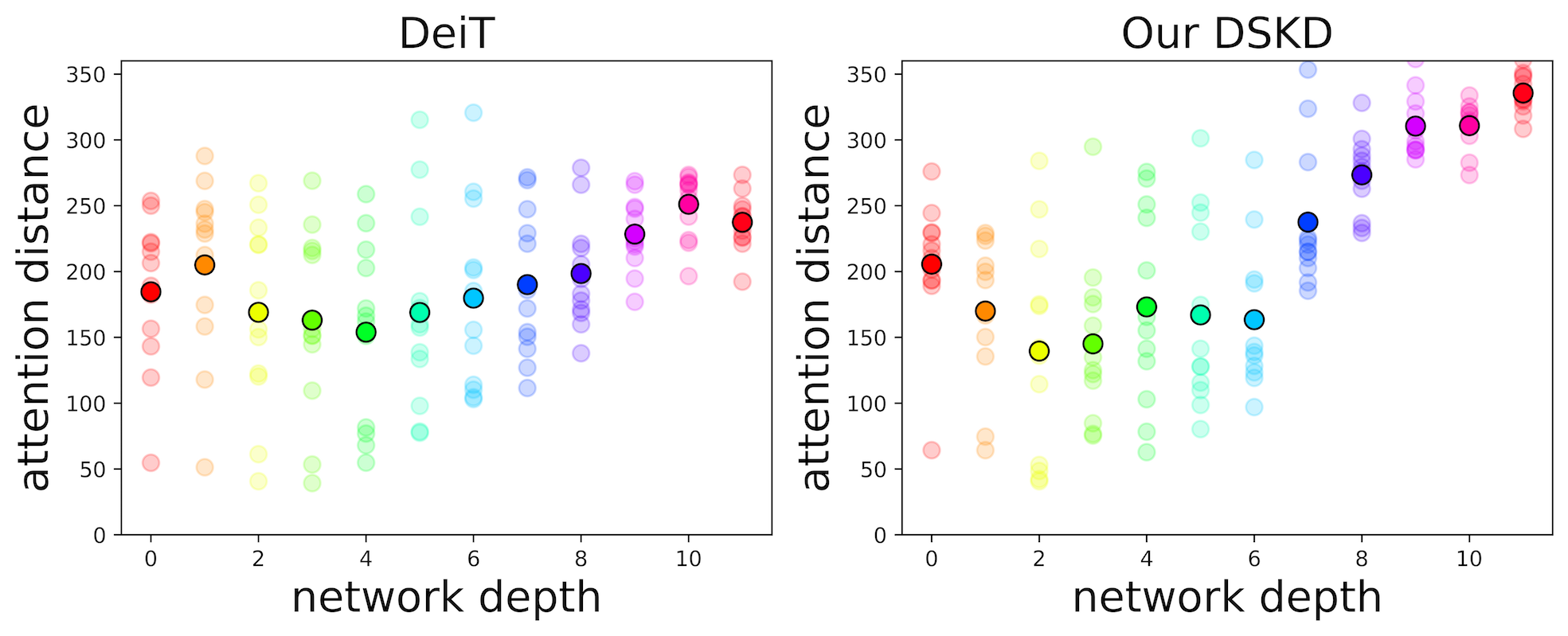}
        \includegraphics[width=0.43\textwidth]{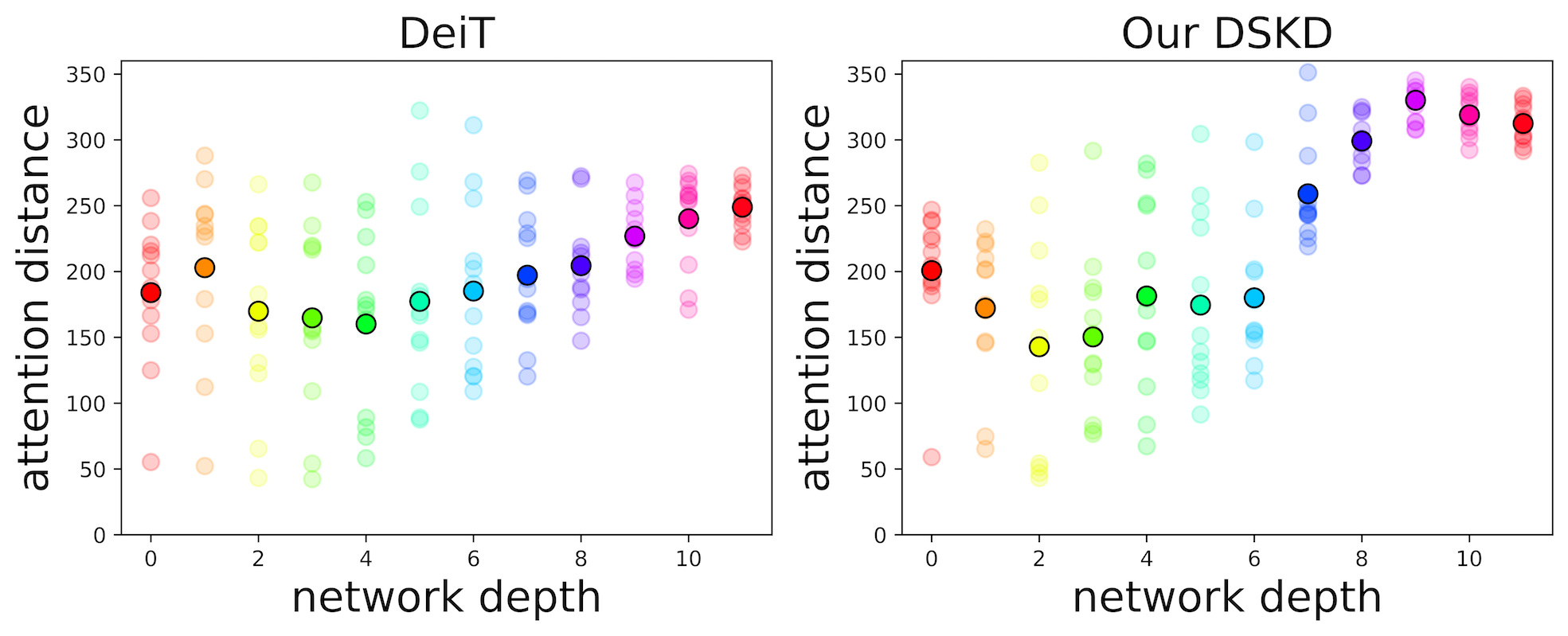}
    \qquad
        \includegraphics[width=0.43\textwidth]{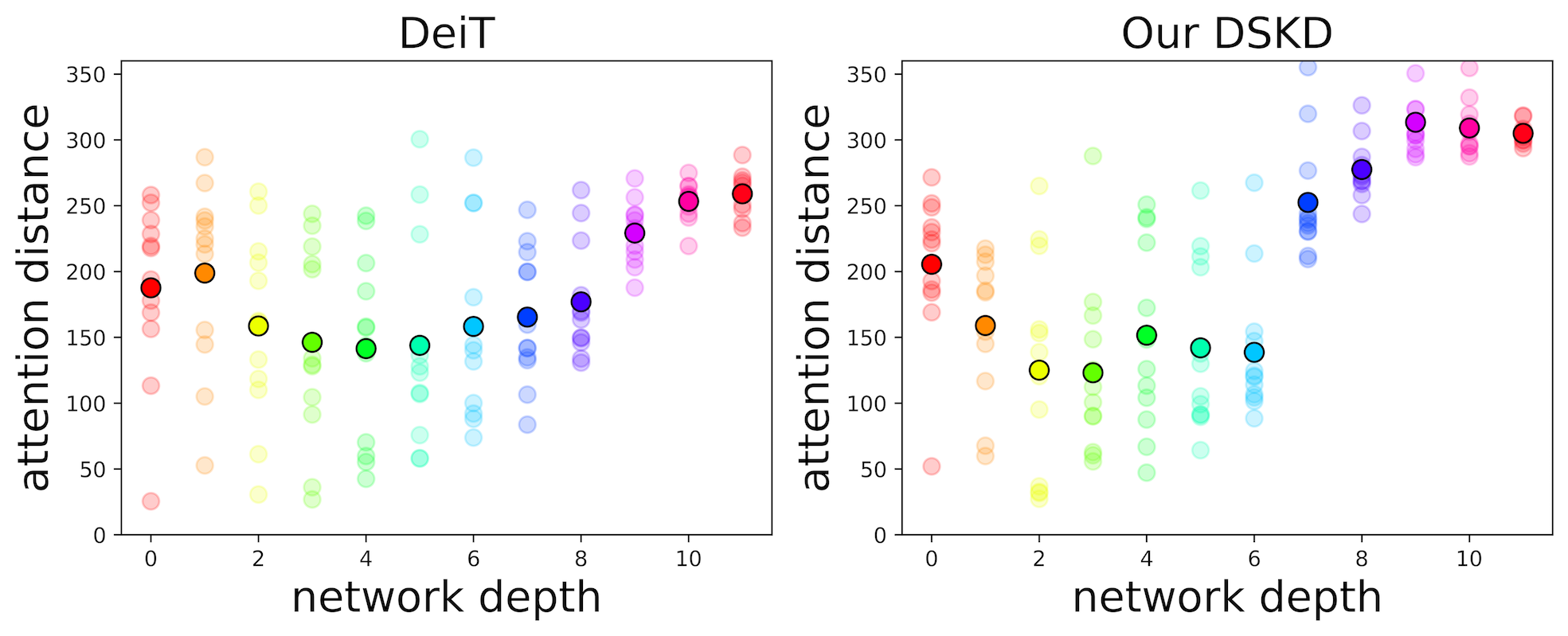}
        \includegraphics[width=0.43\textwidth]{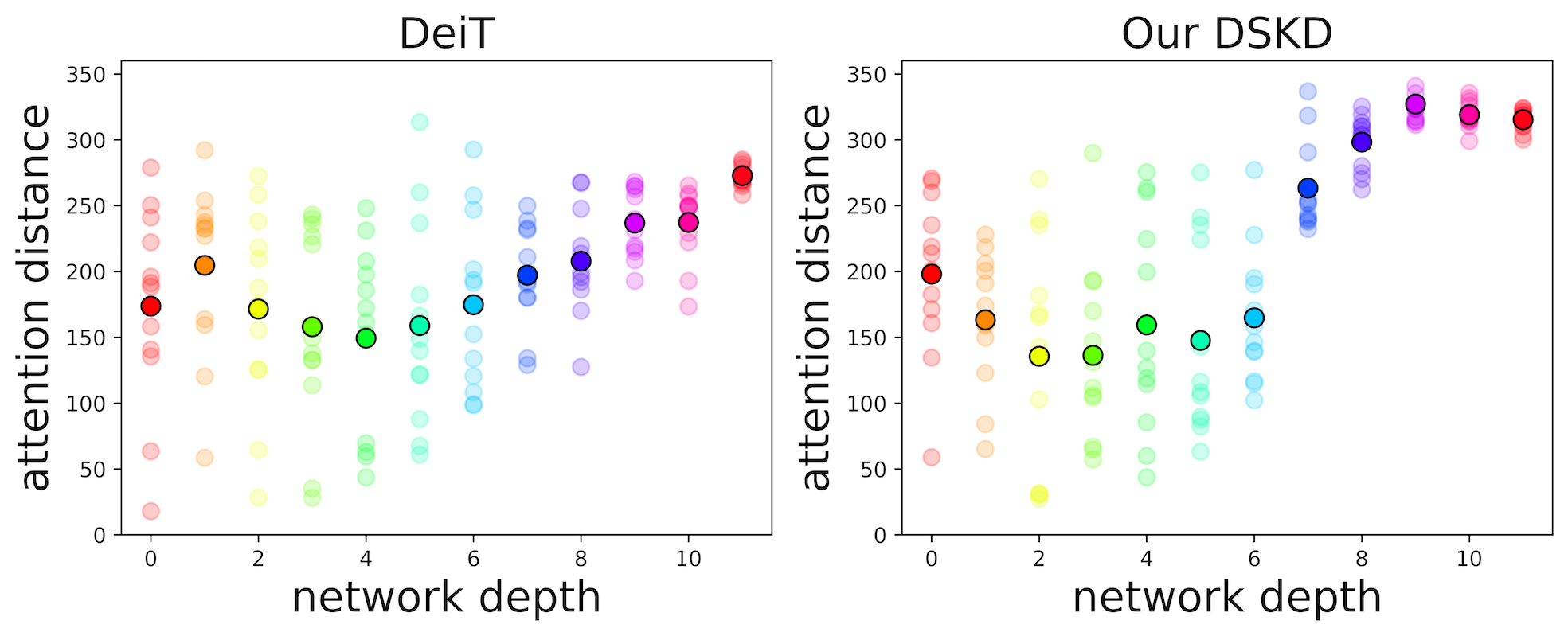}
    
    \caption{\textbf{Mean attention distance} of DeiT-B and CSKD-B for downstream tasks. Results indicate that the global capability of CSKD is generalizable and transferable since the attention distance of DeiT-B is always larger.}
    \label{fig:down_attn_dist}
\end{figure*}

\begin{figure*}[t!]
    \centering
	\includegraphics[width=0.33\textwidth]{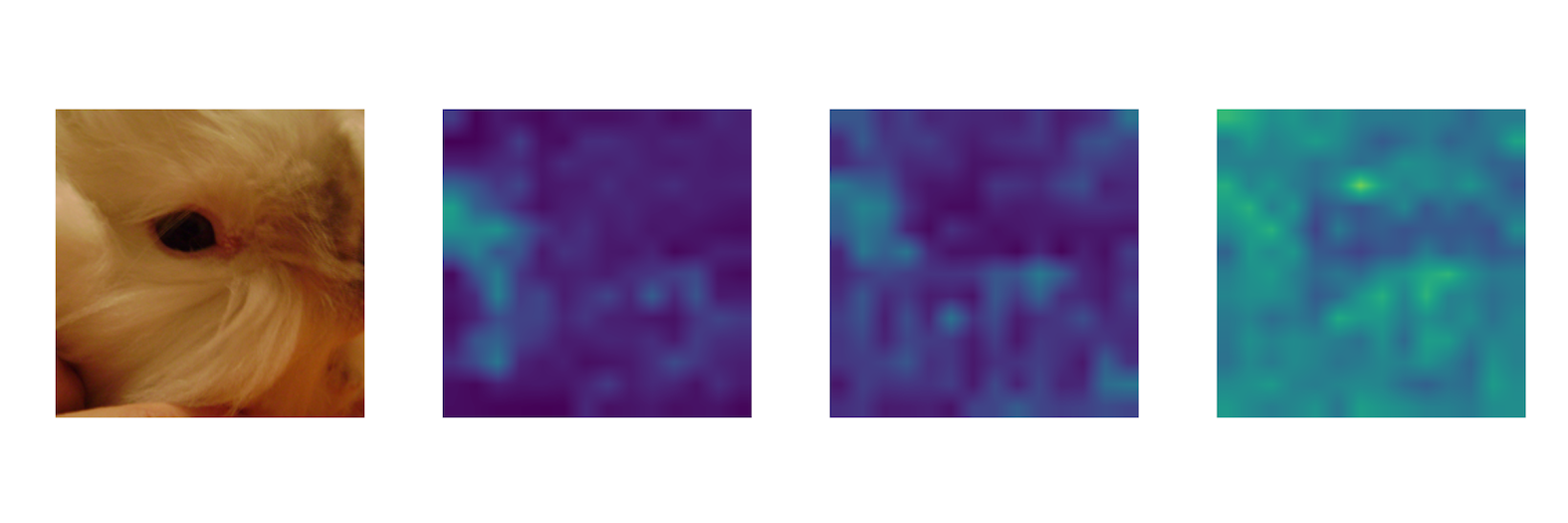}
	\includegraphics[width=0.33\textwidth]{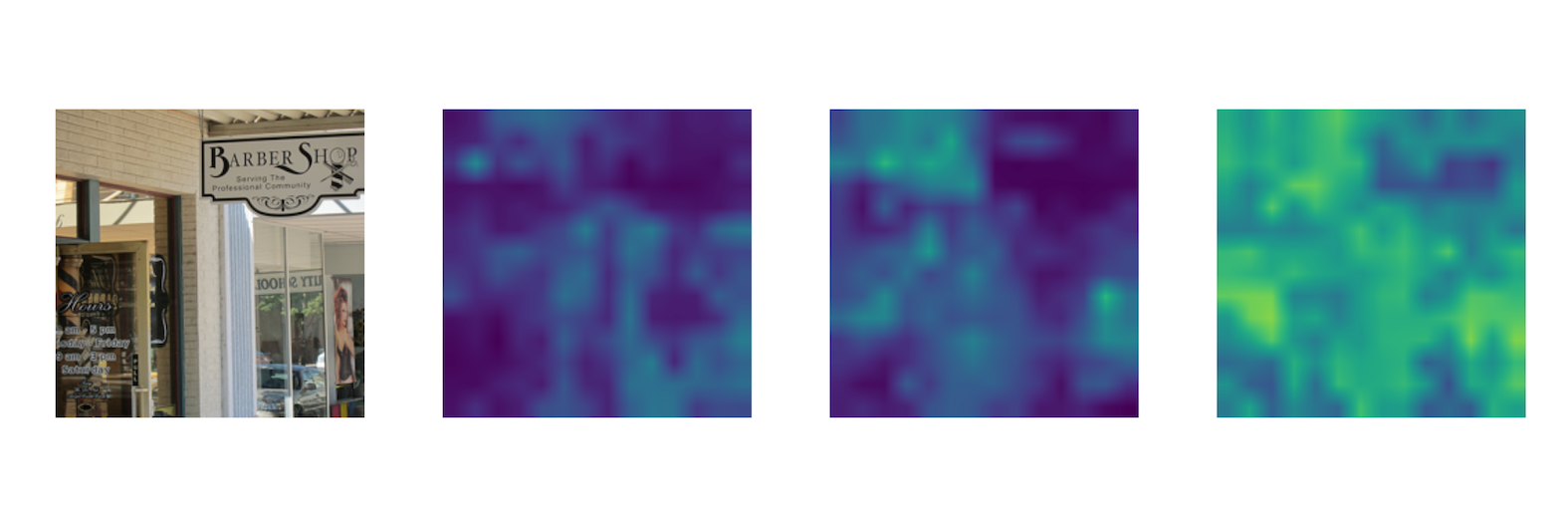}
	\includegraphics[width=0.33\textwidth]{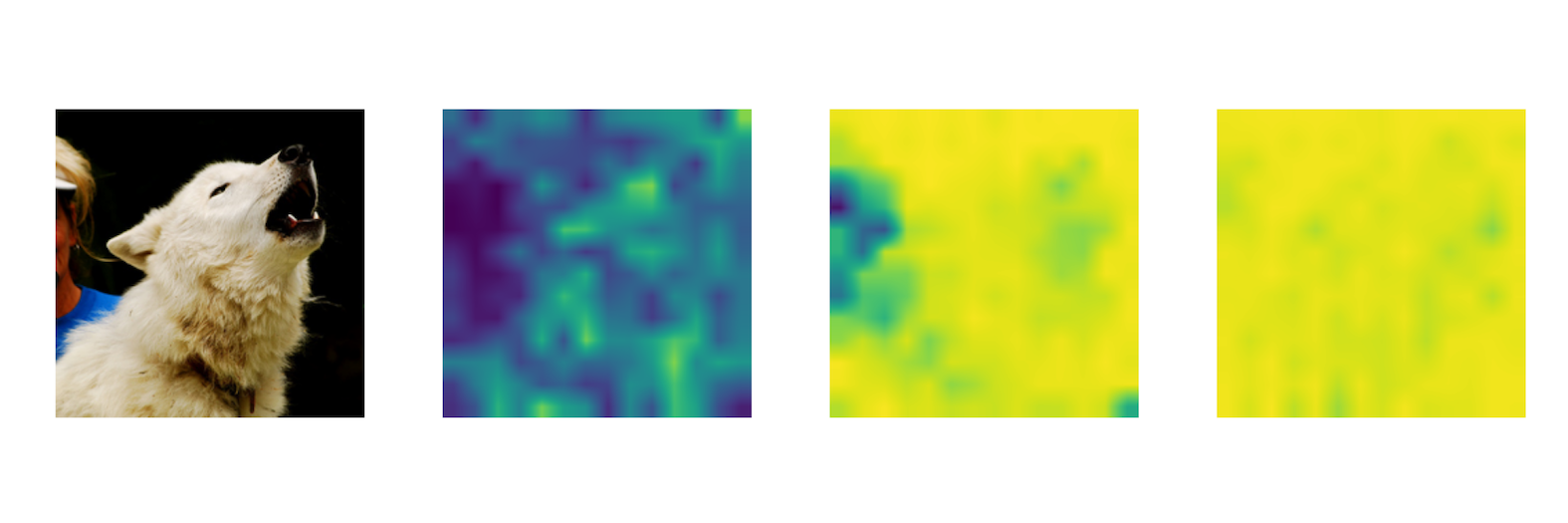}
\includegraphics[width=0.33\textwidth]{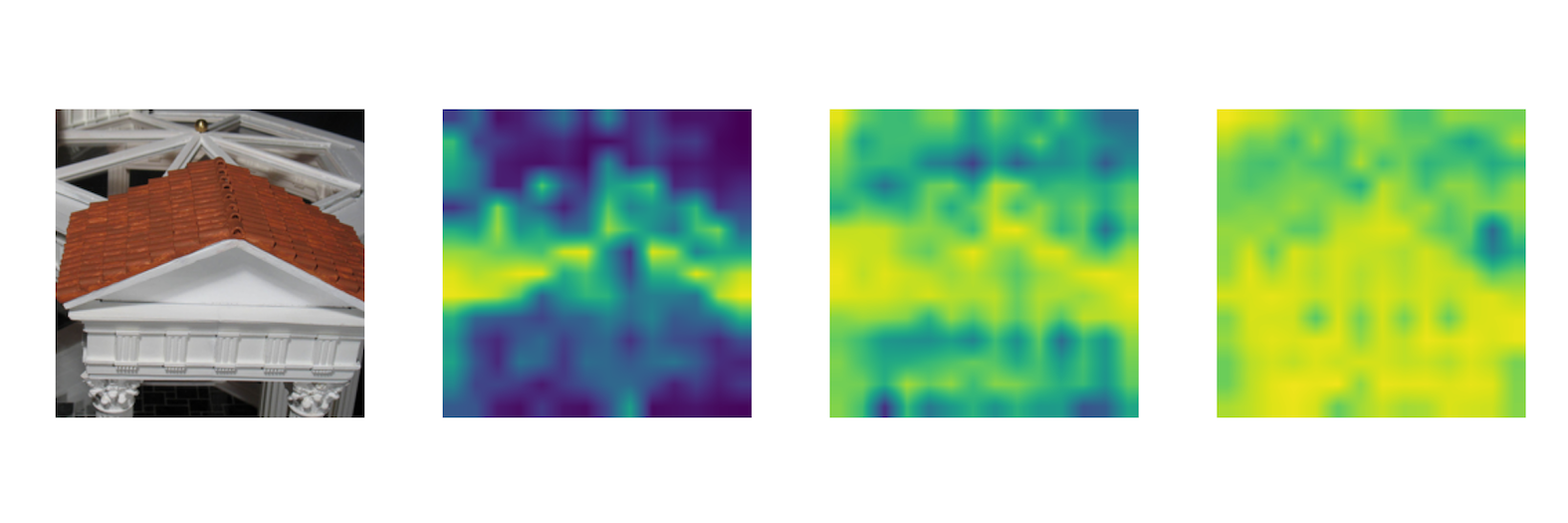}
	\includegraphics[width=0.33\textwidth]{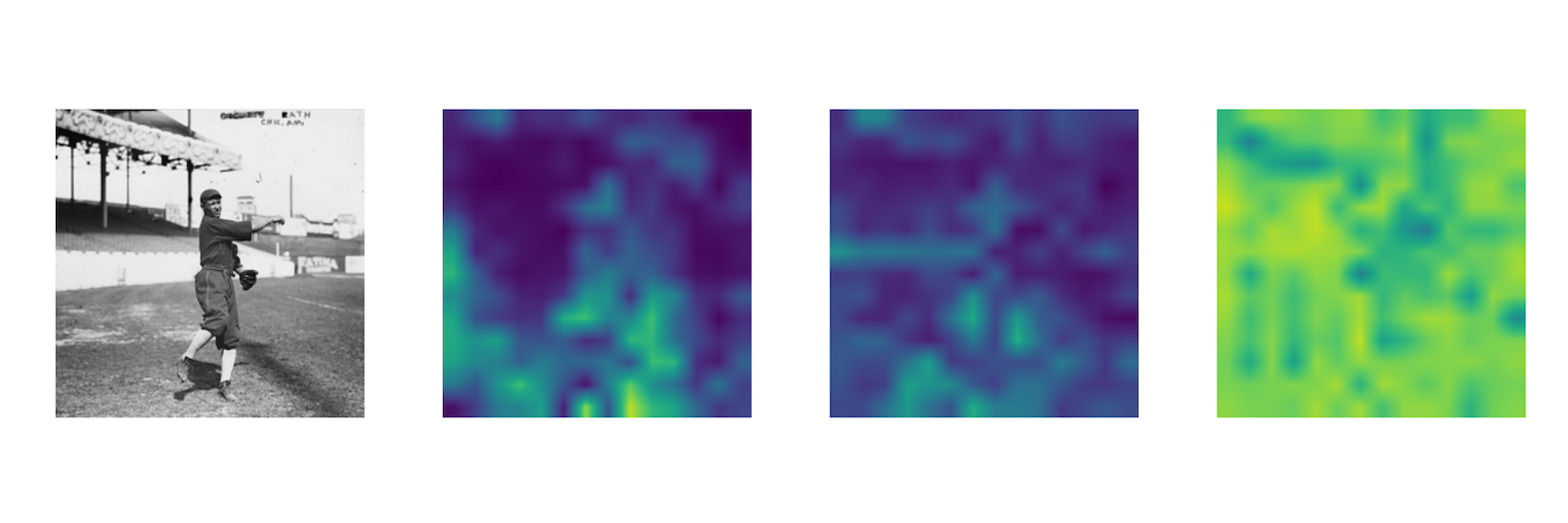}
	\includegraphics[width=0.33\textwidth]{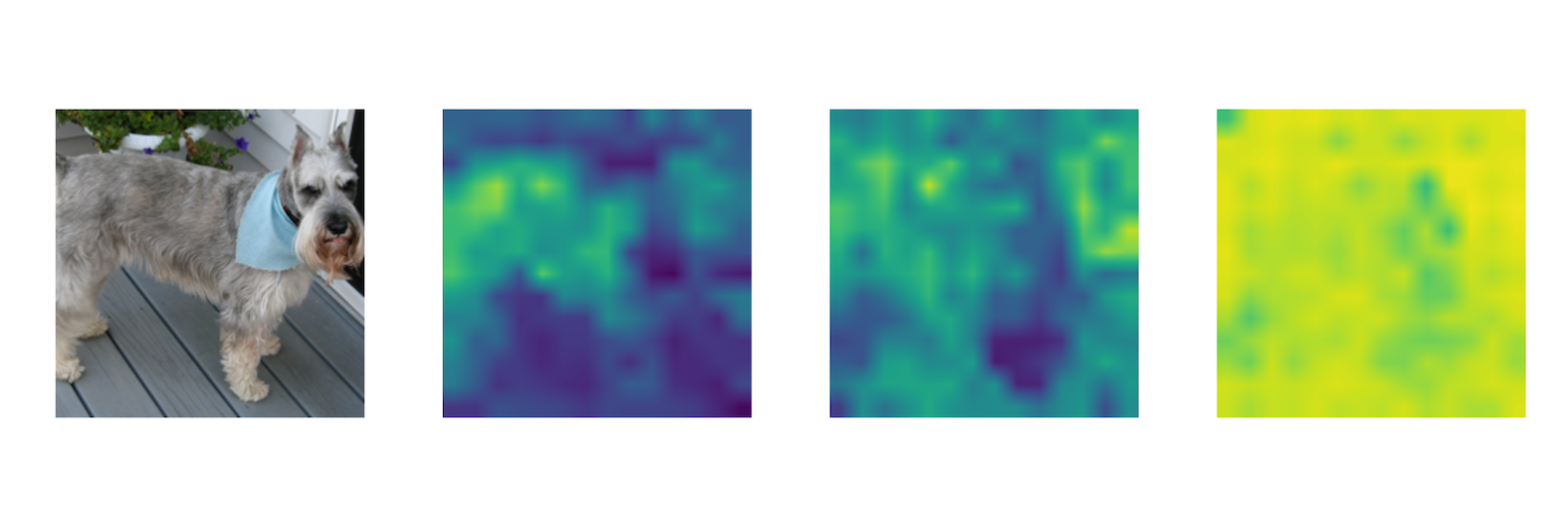}
\includegraphics[width=0.33\textwidth]{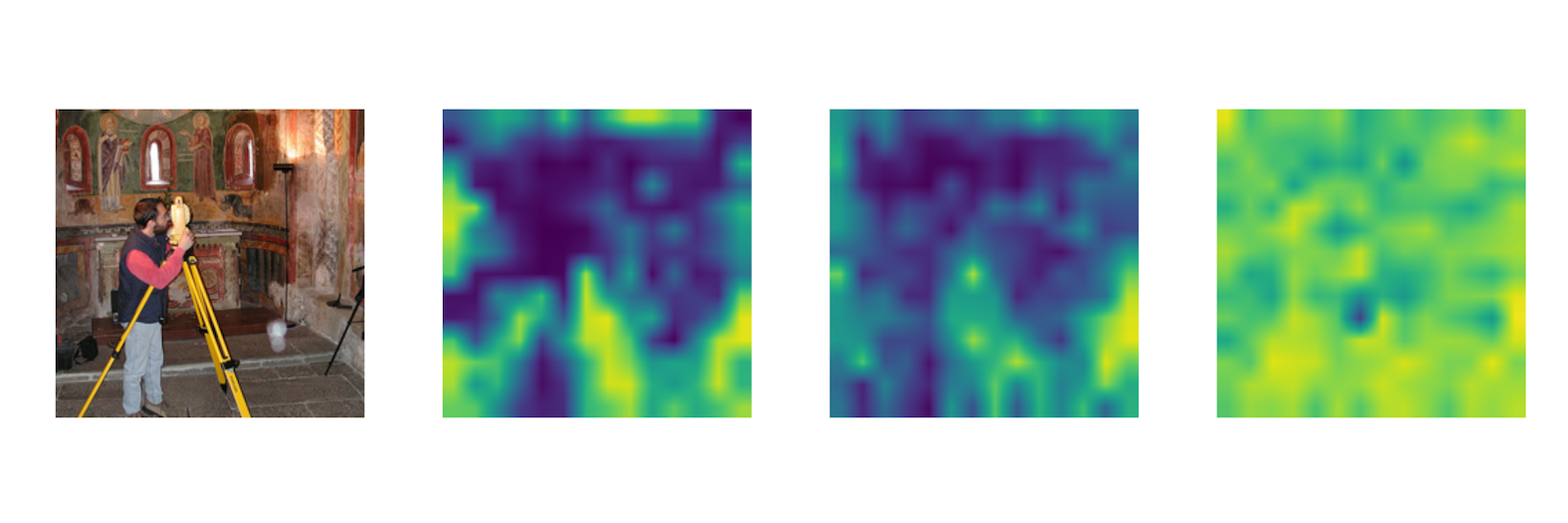}
	\includegraphics[width=0.33\textwidth]{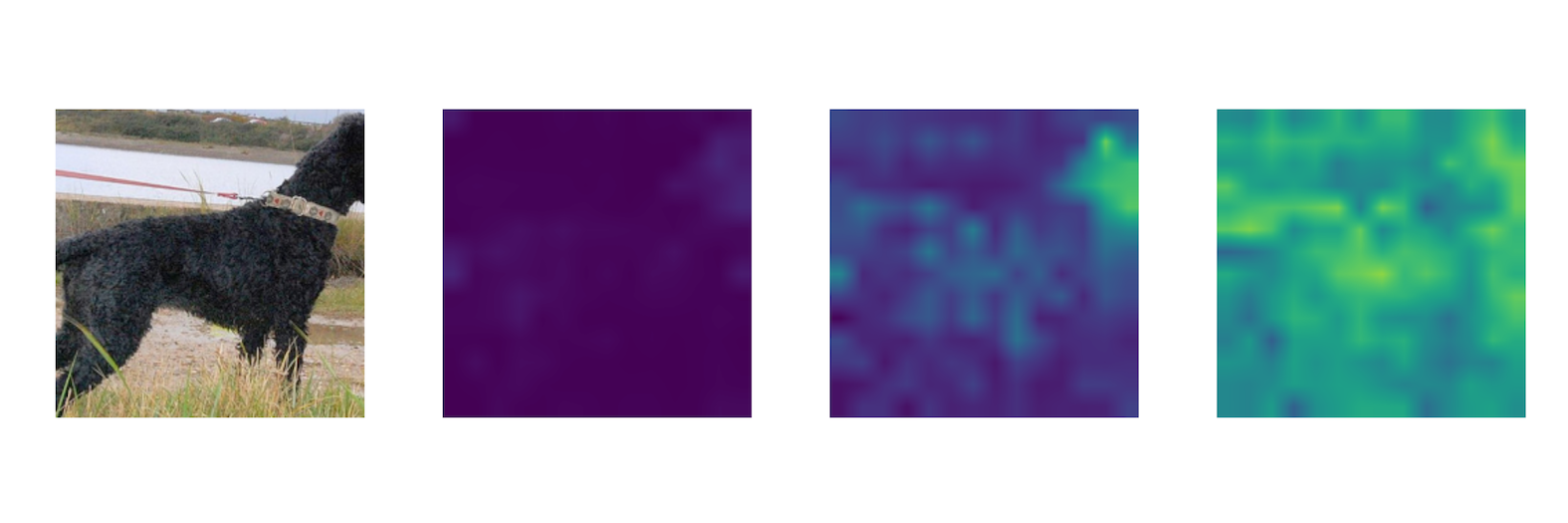}
	\includegraphics[width=0.33\textwidth]{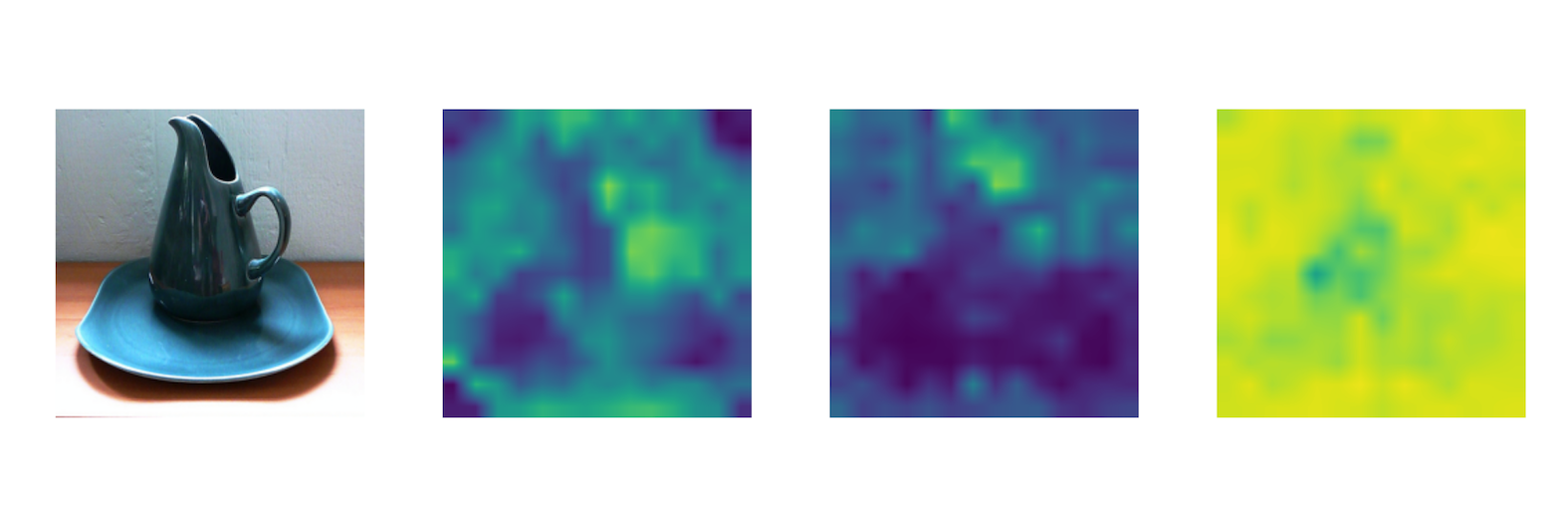}
    \caption{Responses of patch tokens of different epochs. The first column is the input image. The second, third, and fourth ones are response heatmaps of the 100th, 200th, and 300th epochs, respectively.}
    \label{fig:patch_token_response}
\end{figure*}

\subsection{Visualizations and Analysis}
\label{subsec:vis}

In this part, we provide more analysis for a better understanding of CSKD's working mechanism. We mainly study attention distance, attention heatmap, and accuracy dynamics. Visualizations demonstrate the effectiveness of our CSKD and the CKF module.

\vspace{15pt}\noindent{\textbf{Attention Distance.}}
Firstly, we visualize the attention distance of each transformer encoder. Following ViT\cite{vit}, the attention distance is computed for 128 example images by averaging the distance between the query pixel and all other pixels weighted by the attention weight. A larger attention distance means that the global capability is better leveraged. As shown in Figure~\ref{fig:attn_dist}, the attention distance of CSKD-B is higher than the baseline, especially for the latter stages. It suggests that CSKD makes ViT \textit{reach its full potential}. We further explore the attention distance for the downstream tasks to validate that the learned global capability is generalizable and transferable in Figure~\ref{fig:down_attn_dist}.

\vspace{15pt}\noindent{\textbf{Accuracy Dynamics.}}
We investigate how CKF influences the training process by visualizing the validation accuracy dynamics. As shown in Figure~\ref{fig:curve}, our CKF module mainly works in the later training period \textit{as designed}. The performance gap becomes larger in the latter epochs. It suggests that our CKF gives full play to ViT's network capability when the network is almost converged.

\vspace{15pt}\noindent{\textbf{Attention Heatmaps.}}
We visualize the attention heatmaps to validate the superiority of our CSKD. Visualizations in Figure~\ref{fig:attnmap} demonstrate that the attention heatmaps of CSKD are better than the baseline, \ie, CSKD focuses more attention on the salient object of the image.

\vspace{15pt}\noindent{\textbf{Responses Dynamics of Patch Tokens.}}
We visualize the response dynamics of patch tokens during the training process. As shown in Figure~\ref{fig:patch_token_response}, CKF helps patch tokens predict more accurate labels with the progress of training, \ie, the global relation is modeled better and better.

\section{Conclusion}
Distilling knowledge from CNN to ViT achieves great success. In this paper, we reveal two problems limiting the distillation performance. On one hand, spatial-wise knowledge transfer is inefficient since the features are hard to align between CNN and ViT. On the other hand, ViT's network capability is limited in the later training period. To address these problems, we present Cumulative Spatial Knowledge Distillation~(CSKD). CSKD distills knowledge to patch tokens from the corresponding local responses of CNN, transferring spatial-wise knowledge without introducing intermediate features. Moreover, CSKD introduces a Cumulative Knowledge Fusion~(CKF) module, which benefits ViT from the local-inductive-bias knowledge in the early period and gives full play to ViT's capability later. Experiments and analysis on ImageNet-1k and downstream tasks demonstrate the superiority of CSKD.

{\small
\bibliographystyle{ieee_fullname}
\bibliography{egbib}
}

\clearpage
\section*{A.Appendix}
\subsection*{A.1 Details about aligning ViT's dense predictions to CNN's dense predictions.}
In this part, we provide the details about aligning dense predictions. The shape for ViT's dense predictions is $(1, N, C)$; and the shape of CNN's dense predictions is $(1, H, W, C)$. Classical CNNs have a downsampling ratio of 32 while ViT takes $16\times16$ sub-images as patches. Thus, the number of ViT tokens is 4 times the number of CNN's dense responses~($N = 2H\times 2W$). Given a $224\times 224$ image, the last feature map of a classicial CNN is in shape $7\times7$ and the number of patch tokens is 196~($14\times 14$). Hence, it's needed to align the spatial dimensions of the two responses. In CSKD, we simply use a $2\times 2$ average pooling~(the stride is set as 2) to downsample ViT's responses. We first reshape ViT's responses to $(1, 14, 14, C)$, then feed them to the pooling layer to generate the aligned responses with the shape of $(1, 7, 7, C)$. Thus, The shapes of CNN's dense predictions and ViT's dense predictions are \textit{exactly the same}. We further calculate the KL-Divergence or cross-entropy loss on the aligned responses.

\subsection*{A.2 Ensemble of patch tokens.}
Since CSKD also trains the patch tokens, it is convenient to utilize them for classification ensemble. In DeiT, the output of distillation token is combined with the output of class token for the final prediction. In CSKD, the outputs of all patch tokens are further combined. Concretely, we calculate the prediction results of all tokens. All patch token outputs are averaged for the final ``patch token output". Then, we average the ``patch token output", class token output, and distillation token output as the final prediction. Experiments show that it can further improve the performance by 0.1\%. It indicates that the patch tokens are useful for more accurate classification, but the performance is still excellent without extra computation cost~(\ie, without ensemble).

\begin{table}[h]
    \centering
    \begin{tabular}{c|c|c}
        \space & ensemble & top-1 \\
        \Xhline{3\arrayrulewidth}
        \multirow{2}{*}{CSKD-Ti} & \space & 76.1 \\
         & \checkmark & 76.3 \\
        \hline
        \multirow{2}{*}{CSKD-S} & \space & 82.2 \\
         & \checkmark & 82.3 \\
        \hline
        \multirow{2}{*}{CSKD-B} & \space & 83.7 \\
         & \checkmark & 83.8 \\
    \end{tabular}
    \caption{Caption}
    \label{tab:my_label}
\end{table}

\subsection*{A.3 Implementations of transfer learning.}
We follow the training setting of DeiT's repository\footnote{\href{https://github.com/facebookresearch/deit/issues/105}{https://github.com/facebookresearch/deit/issues/105}}. We fine-tune all models with $384\times 384$ resolution on the downstream datasets~(without high-resolution fine-tuning on ImageNet-1k).

\subsection*{A.4 More visualizations.}
We provide more visualizations of the attention heatmaps.

\begin{figure}[ht]
\centering
\includegraphics[width=0.22\textwidth]{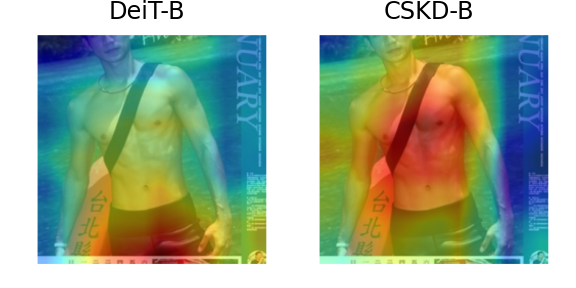}
\includegraphics[width=0.22\textwidth]{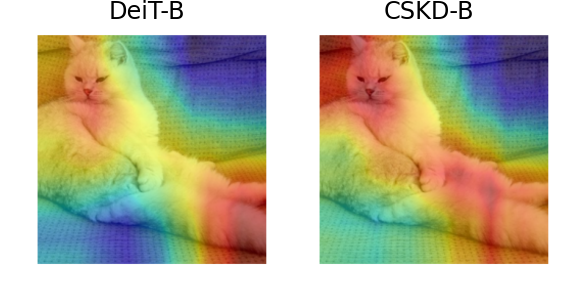}
\includegraphics[width=0.22\textwidth]{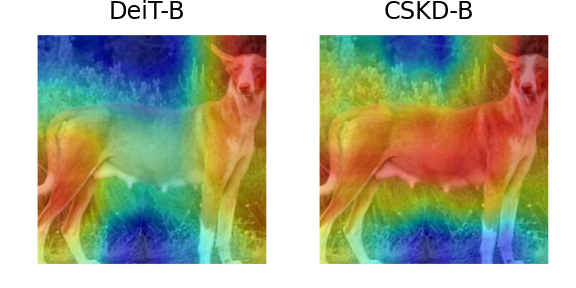}
\includegraphics[width=0.22\textwidth]{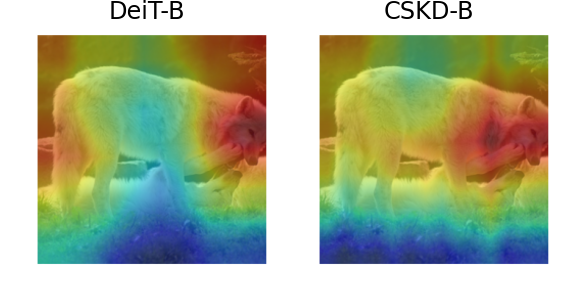}
\quad
\includegraphics[width=0.22\textwidth]{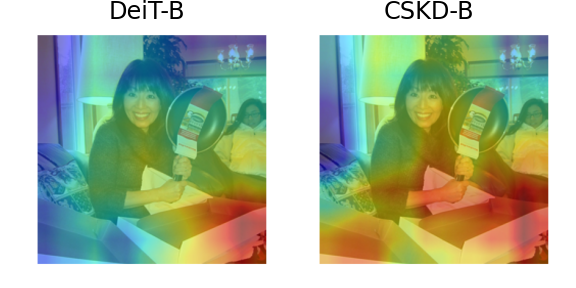}
\includegraphics[width=0.22\textwidth]{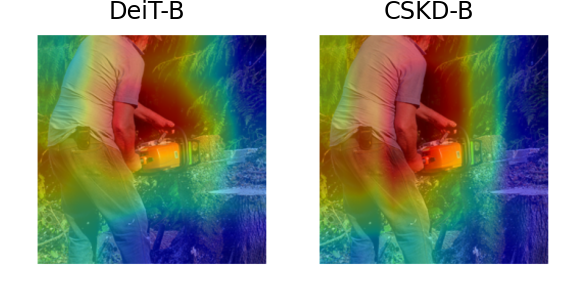}
\includegraphics[width=0.22\textwidth]{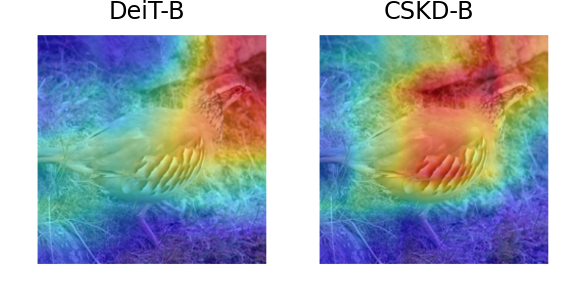}
\includegraphics[width=0.22\textwidth]{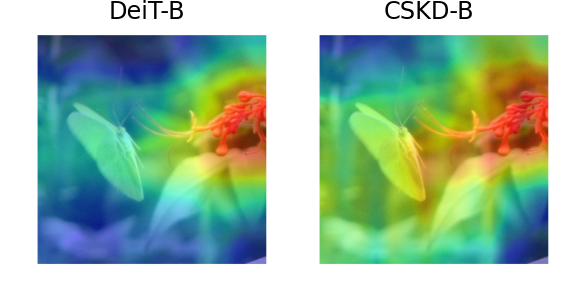}
\quad
\includegraphics[width=0.22\textwidth]{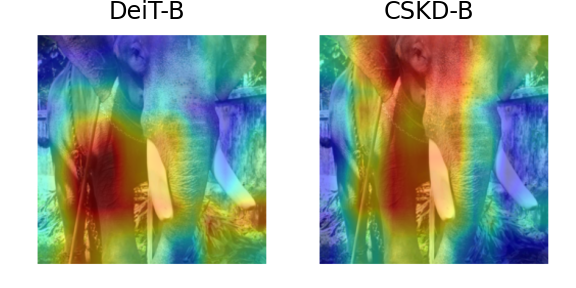}
\includegraphics[width=0.22\textwidth]{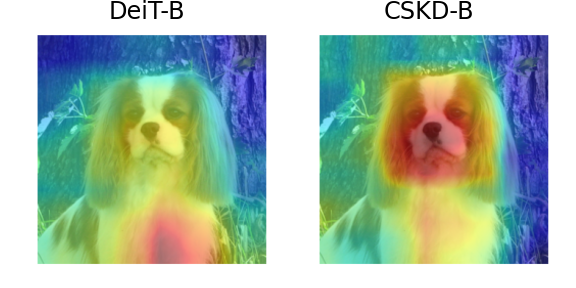}
\includegraphics[width=0.22\textwidth]{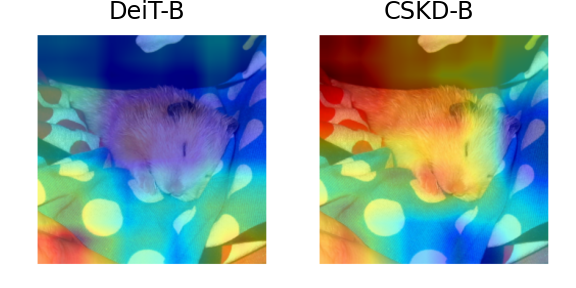}
\includegraphics[width=0.22\textwidth]{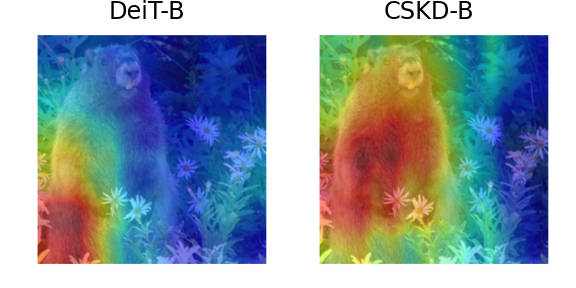}
\quad
\includegraphics[width=0.22\textwidth]{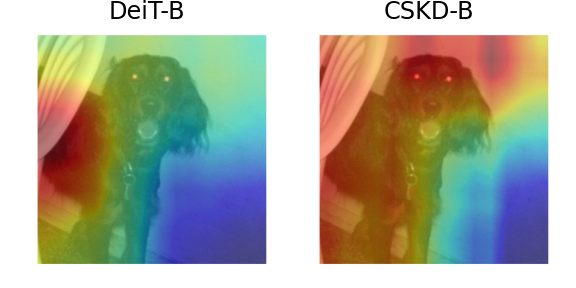}
\includegraphics[width=0.22\textwidth]{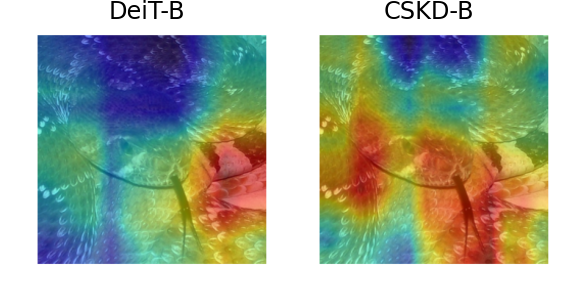}
\includegraphics[width=0.22\textwidth]{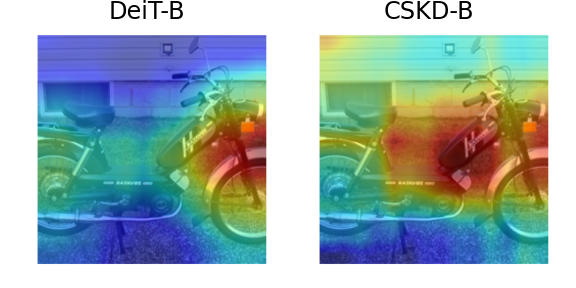}
\includegraphics[width=0.22\textwidth]{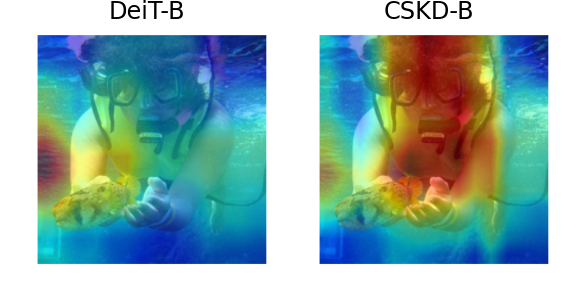}
\caption{Attention Heatmaps from DeiT-B and our CSKD-B. CSKD-B focuses more attention on the salient object.}
\label{fig:supp_attnmap}
\end{figure}

\end{document}